\documentclass[10pt,twocolumn,letterpaper]{article}

\usepackage[pagenumbers]{cvpr}      
\definecolor{cvprblue}{rgb}{0.21,0.49,0.74}
\usepackage[pagebackref,breaklinks,colorlinks,allcolors=cvprblue]{hyperref}

\def\paperID{17431} 
\def\confName{CVPR}
\def\confYear{2026}

\usepackage{booktabs}
\usepackage{multirow}
\usepackage{colortbl}
\usepackage{makecell}
\title{Diffusion As Self-Distillation: End-to-End Latent Diffusion In One Model}

\author{Xiyuan Wang \qquad Muhan Zhang\\
Institute of Artificial Intelligence\\
Peking University\\
{\tt\small \{wangxiyuan, muhan\}@pku.edu.cn}
}

\begin{document}
\maketitle
\begin{abstract}
Standard Latent Diffusion Models rely on a complex, three-part architecture consisting of a separate encoder, decoder, and diffusion network, which are trained in multiple stages. This modular design is computationally inefficient, leads to suboptimal performance, and prevents the unification of diffusion with the single-network architectures common in vision foundation models. Our goal is to unify these three components into a single, end-to-end trainable network. We first demonstrate that a naive joint training approach fails catastrophically due to ``latent collapse'', where the diffusion training objective interferes with the network's ability to learn a good latent representation. We identify the root causes of this instability by drawing a novel analogy between diffusion and self-distillation based unsupervised learning method. Based on this insight, we propose Diffusion as Self-Distillation (DSD), a new framework with key modifications to the training objective that stabilize the latent space. This approach enables, for the first time, the stable end-to-end training of a single network that simultaneously learns to encode, decode, and perform diffusion. DSD achieves outstanding performance on the ImageNet $256\times 256$ conditional generation task: FID=13.44/6.38/4.25 with only 42M/118M/205M parameters and 50 training epochs on ImageNet, without using classifier-free-guidance.
\end{abstract}
\section{Introduction}\label{sec:intro}

\begin{figure*}[t]
\includegraphics[width=\textwidth]{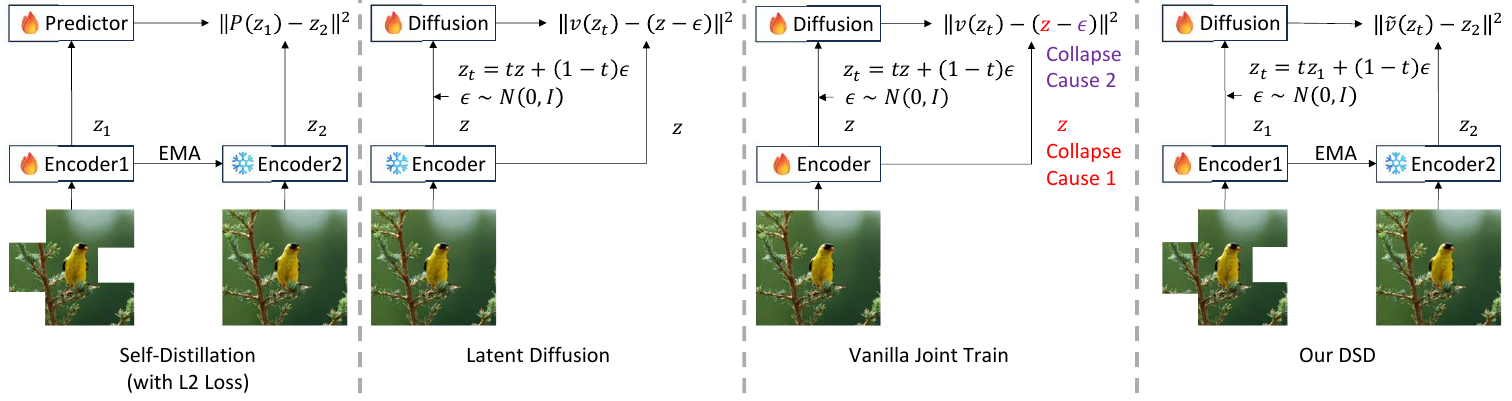}
\caption{
Analogy between Self-Distillation (SD) based unsupervised learning and the standard diffusion model loss, comparing four architectural paradigms. (1)  SD: The Online Encoder ($E_1$) generates a latent $\mathbf{z}_1$ from an augmented image ($\mathbf{x}^{+}$, e.g., with a random mask). $\mathbf{z}_1$ is then passed through a predictor ($P$) and trained to match the target latent $\mathbf{z}_2$ provided by the Target Encoder ($E_2$), which is a frozen copy updated via Exponential Moving Average (EMA). SD is a stable method for producing high-quality latent representations. (2)  Latent Diffusion: The encoder ($E$) is frozen, and only the diffusion model ($v_\theta$) is trained. The clean image latent $\mathbf{z}$ is used in two ways: input $\mathbf{z}_t$ for the diffusion model and target $\mathbf{z}-\mathbf{\epsilon}$ in the $\text{L}2$ loss. (3)  Vanilla Joint Training: Unlike LDM, the encoder ($E$) is unfrozen and optimized with the diffusion loss. This setup leads directly to latent collapse and poor image quality. The reasons for this collapse are structurally identified in this figure (indicated by the colored parts). 4.  Our Diffusion As Self-Distillation (DSD): By implementing solutions that fix the identified collapse mechanisms, our DSD framework achieves a unified architecture where the encoder, predictor, and diffusion backbone are all integrated into a single Vision Transformer (ViT) with task-specific heads.}\label{fig:arch}
\end{figure*}

The design of modern image generation models, particularly Diffusion Models, has heavily centered on the choice of the latent space. Initial models~\citep{ddpm,DDIM,NCSM} operated directly in the high-dimensional pixel space. The breakthrough of Latent Diffusion Models (LDMs)~\citep{ldm,emu,sd3,flux} shifted the focus to a more compact, learned latent space provided by a pretrained Variational AutoEncoder (VAE~\citep{vae}). While the VAE component has been continuously refined~\citep{1dtok, modeseektok, detok, VAVAE}, the standard LDM pipeline remains modular, requiring three distinct, often large, components: a pre-trained Encoder, a pre-trained Decoder, and the main Diffusion Model. This modular design introduces several significant problems:
\begin{itemize}
    \item Lack of Unification: The three-component architecture prevents LDMs from being unified with other major unsupervised learning paradigms, which typically use a single core model. Methods like contrastive learning~\citep{simclr, moco} and self-distillation~\citep{JEPA,dino,BYOL} form the basis of vision foundation models using a single Transformer~\citep{transformer} or CNN~\citep{CNN}. The multi-component nature of LDMs makes them an outlier, hindering their adoption as a standard pre-training task for unified foundation models.
    \item Suboptimal Performance: The multi-stage optimization is suboptimal. The VAE is trained with a reconstruction loss, which does not consider the specific properties required by the downstream diffusion model. On the other hand, most current research focuses on optimizing only one component of the LDM pipeline, such as the autoencoder~\citep{VAVAE,1dtok,detok} or the diffusion model~\citep{fasterdit,sit,repa}, rather than the entire system.
    \item Computational Overhead: Encoder and decoder consume a significant portion of resources. For example, in LDM architectures, the VAE can account for up to 20\% of the total parameters and introduce substantial latency~\citep{RAE}.
\end{itemize}
This motivates our central problem: \textit{Can we unify the encoder, decoder, and diffusion model into a single, end-to-end trainable network?} This unification would drastically simplify the generation pipeline and bridge the gap between latent diffusion and other unsupervised learning methods.

A straightforward method is to simply combine the encoder, decoder, and diffusion model into a single network and train the reconstruction and diffusion tasks simultaneously. However, this attempt fails, as we find in our experiments that the reconstruction loss hardly reduces (as shown in the first image of our \cref{fig:erplot}), resulting in poor reconstruction and generation quality. Previous work REPA-E~\citep{REPA-E} found a similar problem and addressed it by simply preventing the diffusion loss's gradient from backpropagating to the VAE, thereby removing the diffusion loss's impact on the reconstruction task. However, this does not truly solve our problem, as their method detaches the diffusion loss's gradient on the latent, preventing the diffusion task from guiding the latent representation learning. 

In this paper, we first reframe our central problem from an unsupervised learning perspective. As shown in \cref{fig:arch}, the latent diffusion model is similar to self-distillation (SD) based unsupervised learning methods~\citep{BYOL}. Self-distillation method matches the representations of different augmentations of the same data.  They both use an encoder whose latent output is used in two branches: one is used directly in the loss, while the other is transformed by a predictor (or the diffusion model) before being used in the loss. Both can use an L2 loss as the objective. Moreover, a major concern that plagues optimization in such unsupervised methods can be used to describe our observed problem: the reconstruction loss failing to decrease during joint training is a symptom of \textit{collapse}—a rapid reduction in the effective dimensionality of the latent space. As shown in the first image of our \cref{fig:erplot}, the diffusion loss also leads to a rapid decrease in the latent space's effective dimension during training.

Therefore, based on previous theoretical analyses of collapse in unsupervised learning, we provide a systematic and quantitative analysis by identifying two root causes tied directly to the structure of the standard diffusion loss (L2 loss fitting the diffusion velocity):
\begin{itemize}
    \item  Root Cause 1: Latent Variance Suppression. The L2 loss term implicitly contains a term proportional to the variance of the latent representation. Its gradient compels the encoder to minimize latent variance, forcing the latent vectors to cluster around the mean and causing collapse.
    \item  Root Cause 2: Failure of Rank Differentiation. Previous work~\cite{RankDiff} shows that self-distillation relies on a rank differentiation mechanism to avoid collapse, which requires the predictor to reduce the effective dimension of its output features. However, the random noise in the standard diffusion objective is effectively high-rank, which forces the predictor (the diffusion model) to output high-rank signals, violating the SD stability condition.
\end{itemize}
To counteract these specific causes of collapse, we propose the Diffusion as Self-Distillation (DSD) framework, based on two technical innovations:
\begin{itemize}
\item  Decoupling (Addressing Root Cause 1): We use the stop-gradient ($\text{sg}$) operator on the target clean latent to eliminate the gradient path that penalizes latent variance, thereby protecting latent's expressiveness.
\item  Loss Transformation (Addressing Root Cause 2): We analytically prove that the velocity prediction loss is mathematically equivalent to a loss where the model predicts the clean latent. This transformation is key: by targeting the denoised image latent instead of the noisy velocity, the predictor is forced to reduce the effective rank (i.e., act as a denoiser), successfully activating the stable rank differentiation mechanism of self-distillation.
\end{itemize}
With these insights, we achieve a stable, end-to-end training procedure for a single network that simultaneously performs encoding, decoding, and diffusion, leading to a new class of efficient and end-to-end generative models. Our DSD achieves outstanding performance on the ImageNet $256\times 256$ conditional generation task: FID=13.44/6.38/4.25 with no classifier-free-guidance (cfg~\citep{cfg}) using only 42M/118M/205M parameters and 50 training epochs on ImageNet. Our DSD outperforms all baselines with similar number of parameters and can even match latent diffusion methods with up to 700M parameters in total, verifying our method's effectiveness.

\section{Related Work}

\subsection{End-to-End Latent Generative Models}

Latent Score-Based Generative Models (LSGM)~\citep{LSGM} explored joint optimization by an explicit entropy regularization term on the latent space to maintain latent variance and avoid collapse. More recently, Representation-Efficient Partial Autoencoders (REPA-E)~\citep{REPA-E} proposed replacing the traditional VAE with a partial autoencoder that uses the diffusion U-Net's intermediate features for reconstruction.

\paragraph{Relationship to REPA-E~\cite{REPA-E}}

Our DSD achieves full architectural unification, merging the encoder, decoder, and diffusion model into a single, monolithic ViT backbone. REPA-E still maintains the encoder, decoder, and diffusion model as separate, distinct modules. Moreover, We establish the theoretical link between the diffusion loss and self-distillation to diagnose and solve latent collapse. REPA-E does not provide this theoretical diagnosis; instead, it relies on directly detach diffusion loss from VAE.

\subsection{Self-Distillation and Rank Differentiation in Unsupervised Learning}

Self-Distillation (SD) has become a dominant paradigm in unsupervised representation learning, with major models including DINO~\citep{dino}, DiNOv2~\citep{dinov2}, DiNOv3~\citep{dinov3}, BYOL~\citep{BYOL}, and I-JEPA~\citep{JEPA}. These models successfully avoid the problem of dimensional collapse without resorting to explicit dispersive losses. Its architecture is shown in \cref{fig:arch}. The stability of these methods can be explained by the rank differentiation mechanism~\citep{RankDiff}. Our work also leverages this theoretical framework.

\section{Preliminary}

\subsection{Diffusion Model}

Throughout this work, we adopt a continuous-time, noise-conditioned latent diffusion framework, specifically utilizing the velocity prediction objective, consistent with modern architectures like DiT~\citep{dit} and LightningDiT~\citep{VAVAE}. Let $t$ be the time step sampled from a distribution $T$ on $[0, 1]$, typically the uniform distribution $U[0, 1]$. Let $\mathbf{z} \in \mathbb{R}^d$ be the clean image latent representation. A noisy latent $\mathbf{z}_t$ is generated by linearly interpolating between $\mathbf{z}$ and a random noise vector $\mathbf{\epsilon} \sim \mathcal{N}(0, \mathbf{I}_d)$:
\begin{equation}
\mathbf{z}_t = t\mathbf{z} + (1-t)\mathbf{\epsilon}.    
\end{equation}
The standard objective is to train the model $v(\mathbf{z}_t, t)$ to predict the velocity vector from $\mathbf{\epsilon}$ to $\mathbf{z}$, which is $\mathbf{z}-\mathbf{\epsilon}$. The mean squared error (MSE) loss is:
\begin{equation}\label{eq:pre_vanilla_loss}
\mathcal{L}_\text{v} = \mathbb{E}_{t\sim T, \mathbf{\epsilon}\sim \mathcal{N}(0, \mathbf{I}_d), \mathbf{z}\sim p_1} \left[\Vert v(\mathbf{z}_t, t) - (\mathbf{z}-\mathbf{\epsilon})\Vert^2\right],
\end{equation}
where $p_1$ is the distribution of the clean image latent.

The conditional distribution of the noisy latent given the clean latent is $p_t(\mathbf{z}_t|\mathbf{z})=\mathcal{N}(\mathbf{z}_t | t\mathbf{z}, (1-t)^2 \mathbf{I}_d)$, the marginal distribution is $p_t(\mathbf{z}_t)=\int \mathrm{d}\mathbf{z} p_t(\mathbf{z}_t|\mathbf{z})p_1(\mathbf{z})$, and the posterior distribution of the clean latent $\mathbf{z}$ given a noisy latent $\mathbf{z}_t$ is $p_{1|t}(\mathbf{z}|\mathbf{z}_t)=\frac{p_1(\mathbf{z})p_t(\mathbf{z}_t|\mathbf{z})}{p_t(\mathbf{z}_t)}$.

\subsection{Rank Differentiation Mechanism}

We analyze dimensional collapse using the Effective Rank ($\text{erank}$~\citep{erank}), which provides a smooth, continuous measure of the dimensionality of a representation matrix. Given a matrix $\mathbf{A}\in \mathbb{R}^{n\times m}$ (e.g., a batch of $n$ latent vectors of dimension $m$) with non-zero singular values $\sigma_1,...,\sigma_l$. We define the probability distribution $p_i=\frac{\sigma_i}{\sum_{j=1}^l\sigma_j}$. The effective rank is defined via the entropy of this distribution:
\begin{equation}
    \text{erank}(\mathbf{A})=\exp\left(-\sum_{i=1}^l p_i\log p_i\right)
\end{equation}
Collapse corresponds to $\text{erank}(\mathbf{A})$ reduction.

In self-distillation (SD), an online encoder $E_1$ with predictor $P$ is trained to match the output of a target encoder $E_2$ updated via Exponential Moving Average (EMA). The SD loss for a pair of augmented data $(\mathbf{x}^+, \mathbf{x})$ (the data augmentation can be color jitter, randon mask, gaussian noise, ... as in DiNOv3~\citep{dinov3}) is:
\begin{equation}\label{eq:pre_sd_loss}
\mathcal{L}_\text{SD} = \text{Distance}(P(E_1(\mathbf{x}^+)), \text{sg}(E_2(\mathbf{x}))),
\end{equation}
where $\text{sg}$ is the stop-gradient operator, $\text{Distance}$ is a distance function like L2 loss depending on the model design, and $P$ is a lightweight projector model.

\citet{RankDiff} propose rank differentiation mechanism. They find an condition for stable SD training: The target encoder's output representation should have a higher effective rank than the projected representation, i.e. 
\begin{equation}\label{equ:rankdiffcond}
    \text{erank}(E_2(\mathbf{x})) > \text{erank}(P(E_1(\mathbf{x}^+))).
\end{equation}
This condition leads to an rank increasing loop: In each step
\begin{itemize}
    \item As $P(E_1(\mathbf{x}^+))$ is trained to fit $E_2(\mathbf{x})$, the projected representation's effective rank increases.
    \item Through backpropagation of projector, the input representation $E_1(\mathbf{x}^+)$'s effective rank also increases. 
    \item Through EMA update, $\text{erank}(E_2(\mathbf{x}))$ also increases. 
\end{itemize}

To fullfil the condition, $P$ needs to act as a low-pass filter, reducing the effective rank of its input. Otherwise, if $\text{erank}(E_2(\mathbf{x})) < \text{erank}(P(E_1(\mathbf{x}^+)))$, the loop may not form, and collapse happens. 

As for why the predictor reduces the effective rank, there is no strict theory to our best knowledge. However, \citet{RankDiff} explains it intuitively: the predictor is trained to predict similar representations from the different representations of various augmentations, thus forming a low-pass filter that reduces the effective rank.

\begin{figure*}
\includegraphics[width=\textwidth]{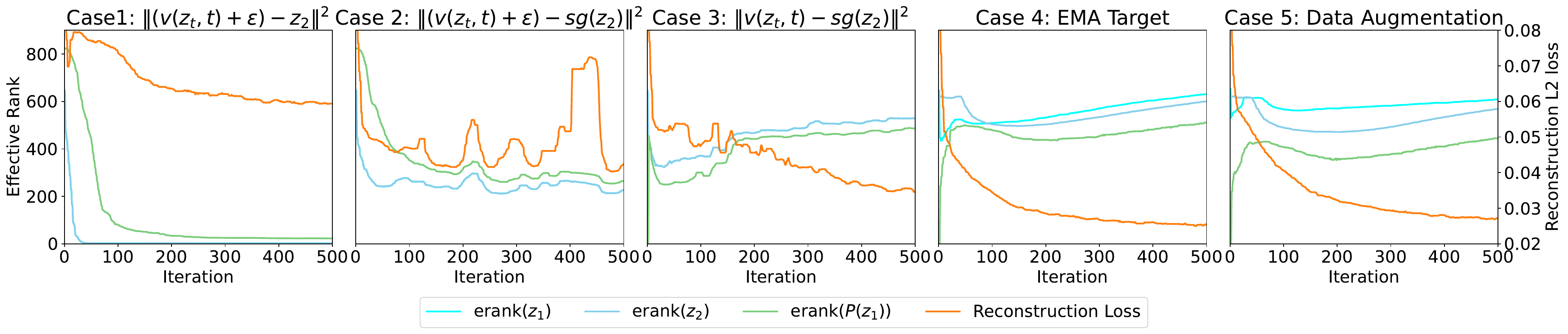}
\caption{Trajectory of latent representations' effective rank and reconstruction loss during end-to-end training process of a ViT-S model on ImageNet. A stable training process is featured with decreasing reconstruction loss (orange lines) and high effective rank (blue lines).}\label{fig:erplot}
\end{figure*}

\section{Diffusion As Self-Distillation (DSD)}

In this section, we provide the step-by-step methodology for constructing our Diffusion as Self-Distillation (DSD) framework. Starting from a naive method for end-to-end latent diffusion and VAE training, we iteratively identify the reasons for collapse and connect the resulting solutions to the self-distillation paradigm.

\subsection{Unified Notation}

To show the similarity between diffusion and self-distillation, we propose a unified notation for them. Both the diffusion loss and the self-distillation loss (with L2 distance)  can be subsumed by the following form:

\begin{equation}
    \Vert P(\mathbf{z}_1, t, \mathbf{\epsilon})- \mathbf{z}_2\Vert^2.
\end{equation}

\begin{itemize}
    \item For self-distillation, $\mathbf{z}_1=E_1(\mathbf{x}^+)$, $\mathbf{z}_2=\text{sg}(E_2(\mathbf{x}))$, the representations for a pair of data from an augmentation strategy, and $P$ is the predictor model with only $\mathbf{z}_1$ as input (i.e., $t$ and $\mathbf{\epsilon}$ are not used).
    \item  For the standard latent diffusion training pipeline, $\mathbf{z}_1$ and $\mathbf{z}_2$ are the same latent representation $\mathbf{z}$ produced by a VAE encoder $E$ from an image. In this case, $\mathbf{z}_1 = \mathbf{z}_2 = \mathbf{z}$, and both are frozen (i.e., no gradients are backpropagated to $E$). And $P$ is much more complex: $P(\mathbf{z}_1, t, \mathbf{\epsilon})=v(z_t, t) + \mathbf{\epsilon}$, with noise $\mathbf{\epsilon}$ sampled from a normal distribution, $t\in [0, 1]$ sampled from a predefined distribution, $z_t=t\mathbf{z}_1+(1-t)\mathbf{\epsilon}$, and $v$ is a diffusion model~\citep{dit}.
    \item For a naive method to train the LDM and VAE end-to-end, $\mathbf{z}_1 = \mathbf{z}_2 = E(\mathbf{x})$, where $E$ is the VAE encoder. These representations require gradients, while $P$ is the same as in the standard LDM pipeline.
    \item For our DSD model, $\mathbf{z}_1 = E_1(\mathbf{x}^+)$ and $\mathbf{z}_2 = \text{sg}(E_2(\mathbf{x}))$, similar to self-distillation. Our predictor $P$ is a Vision Transformer (ViT~\citep{ViT}) $\tilde v$ that takes a noisy version of $\mathbf{z}_1$ as input: $P(\mathbf{z}_1, t, \mathbf{\epsilon}) = \tilde v(z_t, t)$. The model is thus trained to predict the clean target latent $\mathbf{z}_2$.
\end{itemize}

\subsection{Reason 1: Latent Diffusion Loss Contains Latent Variance}

\paragraph{Empirical Evidence (Case 1 in \cref{fig:erplot}).} Directly using diffusion loss $\Vert v(\mathbf{z}_t, t)-(\mathbf{z}_2-\mathbf{\epsilon})\vert^2$ to train encoder makes latent collapse to nearly one single point. 
Case 1 in \cref{fig:erplot} shows a swift, catastrophic drop in the effective rank of the latent representations to near $1$, and the reconstruction loss cannot decrease, confirming the collapse.

\paragraph{Diagnosis: Variance Suppression.}
Similar to the variance-bias decomposition in classic machine learning, the diffusion loss can be decomposed as follows:
\begin{multline}
    \mathbb{E}_{t\sim T}w_t\mathbb E_{\mathbf{z}_t\sim p_t(\cdot)}  \big[\Vert f(\mathbf{z}_t, t) - \mathbb {E}_{\mathbf{z}_2\sim p_{1|t}(\cdot|\mathbf{z}_t)} \mathbf{z}_2\Vert^2
    \\
    + Var_{\mathbf{z}_2\sim p_{1|t}(\cdot|\mathbf{z}_t)} \mathbf{z}_2\big],
\end{multline}
where $w_t\in \mathbb{R}^+$ is positive coefficient, $f(\mathbf{z}_t, t)$ is an auxiliary function containing the original diffusion model $v(\mathbf{z}_t, t)$, and $Var_{\mathbf{z}_2\sim p_{1|t}(\cdot|\mathbf{z}_t)} \mathbf{z}_2$ is the variance of image latent on posterior distribution. The first term, $f(\mathbf{z}_t, t) - \mathbb {E}_{\mathbf{z}_2\sim p_{1|t}(\cdot|\mathbf{z}_t)} z_2$,  drives $f$ to fit the expectation of posterior distribution. The second term only contains $\mathbf{z}_2$, the target representation, and will push the posterior distribution to a single point. Minimizing this term via gradient flow actively forces the encoder $E$ to reduce the spread of the latent space, directly causing collapse.

\paragraph{Solution 1: Decoupling.}
The solution is to eliminate the gradient path through the variance term. We adopt the SD structure by applying a stop-gradient operator ($\text{sg}$) to the target latent $\mathbf{z}_2$, and the loss becomes:
\begin{equation}
\Vert v(\mathbf{z}_t, t) - (sg(\mathbf{z}_2)-\epsilon)\Vert^2
\end{equation}
This operation prevents the diffusion objective from penalizing the latent variance, thereby protecting the expressiveness of the latent space. Note that this decoupling is different from detach in REPA-E~\citep{REPA-E}, as our diffusion loss can still have gradient on latent through $v(\mathbf{z}_t, t)$, where $\mathbf{z}_t$ contains $\mathbf{z}_1$, which is not detached.

\subsection{Reason 2: High-Rank Projection Output of Diffusion Head}

\paragraph{Empirical Evidence (Case 2 in \cref{fig:erplot}).}
As shown in case 2 of \cref{fig:erplot}, when stopping the gradient flow to $\mathbf{z}_2$, the effective rank of latent drops slower, but collapse still happens. Why? We notice that the diffusion loss still affects the encoder through $\mathbf{z}_1$, and the trajectory shows that $\text{erank}(\mathbf{z}_2) < \text{erank}(P(\mathbf{z}_1, t, \mathbf{\epsilon}))$, which violates the rank differentiation mechanism.

\paragraph{Diagnosis: Rank Differentiation Violation.}
Even after applying $\text{sg}(\mathbf{z}_2)$, collapse still occurs, albeit slower (Case 2 in \cref{fig:erplot}). This is because the diffusion loss still affects the encoder via $\mathbf{z}_1$, requiring the rank differentiation mechanism for stability. The core requirement for stability is that $\text{erank}(\mathbf{z}_2) > \text{erank}(P(\mathbf{z}_1, t, \mathbf{\epsilon}))$.

The predictor for latent diffusion $P(\mathbf{z}_1, t, \mathbf{\epsilon})=v(\mathbf{z}_t, t) + \mathbf{\epsilon}$ contains full-rank noise term $\mathbf{\epsilon}$. The noise term makes the output  high-rank, reversing the required inequality: $\text{erank}(\mathbf{z}_2) < \text{erank}(P(\mathbf{z}_1, t, \mathbf{\epsilon}))$. Thus, the rank differentiation mechanism is broken, and collapse happens. A direct solution is to remove the added $\mathbf{\epsilon}$ term, but is it still an diffusion loss? The next paragraph provides a positive answer.

\paragraph{Solution 2: Loss Transformation to Clean Latent Prediction.}

\begin{figure}
\centering
\includegraphics[width=0.6\linewidth]{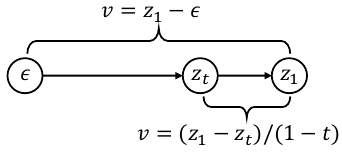}
\caption{Geometric interpretation of the loss transformation. The velocity $\mathbf{v}=\mathbf{z}-\mathbf{\epsilon}$ is proportional to $\mathbf{z}-\mathbf{z}_t$. Therefore, training a model to predict the clean end point $\mathbf{z}$ (denoising) is equivalent to training a model to predict the velocity.}
\label{fig:velotrans}
\end{figure}

We start from the standard diffusion loss:
\begin{equation}\label{eq:v_loss_restate}
\mathcal{L}_{\text{v}} = \mathbb{E}_{t\sim T, \mathbf{z}\sim p_1, \mathbf{\epsilon}\sim \mathcal{N}(0, \mathbf{I}_d)}
\Vert v(\mathbf{z}_t, t) - (\mathbf{z}-\mathbf{\epsilon})\Vert^2.
\end{equation}
As shown in \cref{fig:velotrans}, the velocity of a path can be computed by the start point $\mathbf{\epsilon}$ and end point $z_1$, but can also be computed with the middle point $z_t$ and end point $z_1$ given the time interval. Using the relationship $\mathbf{z}-\mathbf{\epsilon}=\frac{1}{1-t}(\mathbf{z}-\mathbf{z}_t)$, the loss is mathematically equivalent to:
\begin{equation}\label{eq:z_loss_inter}
\mathcal{L}_{\text{v}} = \mathbb{E}_{t\sim T, \mathbf{z}\sim p_1, \mathbf{z}_t\sim p_t(\cdot|\mathbf{z})}
\left\Vert v(\mathbf{z}_t, t) - \frac{1}{1-t}(\mathbf{z}-\mathbf{z}_t)\right\Vert^2,
\end{equation}
Now, we define our clean latent prediction head $\tilde{v}$ (the actual projector $P$ in DSD) to estimate $\mathbf{z}$. The relationship between the two prediction output is given by:
\begin{equation}
v(\mathbf{z}_t, t) = \frac{1}{1-t}(\tilde{v}(\mathbf{z}_t, t) - \mathbf{z}_t).
\end{equation}
Substituting this back into the expectation yields the transformed loss ($\mathcal{L}_{\text{z}}$):
\begin{equation}\label{eq:z_loss_final}
\mathcal{L}_{\text{z}} = \mathbb{E}_{t\sim T}
w_t
\mathbb{E}_{\mathbf{z}\sim p_1, \mathbf{z}_t\sim p_t(\cdot|\mathbf{z})}
\Vert \tilde{v}(\mathbf{z}_t, t) -\mathbf{z}\Vert^2,
\end{equation}
where $w_t=(1-t)^{-2}$ is the time-dependent weight.

By training $\tilde{v}$ to minimize $\mathcal{L}_{\text{z}}$, the predictor is forced to fit the expectation of the clean image latent $\mathbf{z}$. This denoising task forces $\tilde{v}$ to act as a low-pass filter, successfully implementing the required SD stability mechanism.

\paragraph{Empirical Evidence (Case 3 in \cref{fig:erplot}).}
The third image in \cref{fig:erplot} (Decoupling + Loss Transformation) demonstrates successful stability: the reconstruction loss drops successfully, and the representation's effective rank increases smoothly, confirming that the transformation is key to stability. 

\paragraph{Implementation Note:}
For generative sampling, the velocity $v$ must be recovered. We can calculate $v(\mathbf{z}_t, t) = \frac{1}{1-t}(\tilde{v}(\mathbf{z}_t, t)-\mathbf{z}_t)$. However, due to numerical instability from the $\frac{1}{1-t}$ factor when $t \to 1$ (near the end of the diffusion process). To solve this problem, we use a separate and gradient-detached diffusion head $v$. It takes a detached representation (while already contains information for $v$ due to training of $\tilde v$) as input and trained with the standard diffusion noise $\Vert v(\mathbf{z}_t, t)- (\text{sg}(z)-\epsilon)\Vert ^2$

\subsection{DSD Enhancements: Full Self-Distillation Bridge}

The stable training (Case 3) can be further improved by fully adopting SD practices.

\paragraph{EMA Update for Target Encoder ($E_2$).}
In Case 3 of \cref{fig:erplot}, the training is unsmooth because the online encoder $E_1$ acts as both the source and the target. The rapidly changing target $\mathbf{z}_2$ makes prediction difficult. We introduce an EMA-updated target encoder $E_2$ to provide a temporally smoothed, high-quality target $\mathbf{z}_2$.

\paragraph{Empirical Evidence (Case 4 in \cref{fig:erplot}).}
The fourth image in \cref{fig:erplot} (EMA Target) shows a significantly smoother loss curve and faster decrease, confirming the benefit of the EMA update.

\paragraph{Data Augmentation for Online Latent ($\mathbf{z}_1$).}
To strengthen the low-pass filter property of the projector $\tilde{v}$, we apply aggressive data augmentation (masking, Gaussian blur, color jittering) to the input image $\mathbf{x}$ to obtain $\mathbf{x}^+$. This forces the predictor to map a wider variety of noisy, augmented latent inputs ($\mathbf{z}_t$ derived from $\mathbf{x}^+$) to the consistent, clean target $\mathbf{z}_2$.

\paragraph{Empirical Evidence (Case 5 in \cref{fig:erplot}).}
The fifth image in \cref{fig:erplot} (Augmentation) shows that a high mask ratio (0.75) for the input $\mathbf{x}^+$ causes a significant drop in the effective rank of the projected representation $\tilde{v}(\mathbf{z}_t)$, making the stability inequality $\text{erank}(\mathbf{z}_2) > \text{erank}(\tilde{v}(\mathbf{z}_t))$ hold more robustly and ensuring smoother SD training. We combine a variety of augmentations, including color jittering, masking, solarization..., following DINOv3~\citep{dinov3}.

\subsection{Unified Architecture and Loss Objectives}

\begin{figure}
\centering
\includegraphics[width=1.0\linewidth]{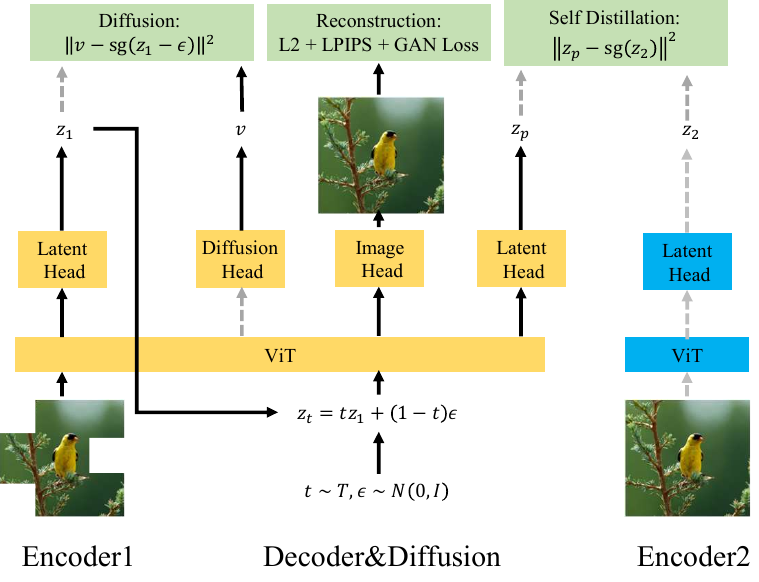}
\caption{Our Unified DSD Architecture. Different ViT block in the figure are essentially one single ViT backbone with different heads. The green block are three objectives in DSD. The yellow blocks are modules updated by gradient descent. The blue blocks are updated by EMA. The black lines denote ordinary forward process. The grey dot line are forward process with stop gradient.}
\label{fig:arch2}
\end{figure}

Our final DSD method integrates all components into a single Vision Transformer (ViT~\citep{ViT}) backbone with multiple task-specific heads, as shown in \cref{fig:arch2}. 

Following our previous analysis, to implement diffusion, we use self-Distillation Loss and Detached Velocity Loss.

\paragraph{Self-Distillation Loss.}
This loss drives the representation learning for the ViT backbone ($E_1$):
\begin{equation}
\mathcal{L}_{\text{DSD}} = \mathbb{E}_{t, \mathbf{x}, \mathbf{x}^+} \Vert \tilde{v}(\mathbf{z}_t, t) - \text{sg}(z_2)\Vert^2,
\end{equation}
where $\mathbf{z}_t=t\mathbf{z}_1+(1-t)\mathbf{\epsilon}, \mathbf{z}_1=E_1(\mathbf{x}^+), \mathbf{z}_2=E_2(\mathbf{x})$. Note that due to numerical stability, we remove the weight $w_t=(1-t)^{-2}$ for different time. 

\paragraph{Detached Velocity Loss.}
A separate head $v$ is trained for the final sampling process. Crucially, the gradient for the target velocity is stopped to ensure it does not corrupt the DSD-trained backbone features:
\begin{equation}
\mathcal{L}_{velo} = \mathbb{E}_{t, \mathbf{z}, \mathbf{\epsilon}} \left[\Vert v(\mathbf{z}_t, t) - \text{sg}(\mathbf{z}-\mathbf{\epsilon})\Vert^2\right].
\end{equation}

Furthermore, as a latent diffusion model, DSD should have the capacity to convert image to latent and reverse.
\paragraph{Reconstruction Training.}
The ViT acts as a full autoencoder. The encoder converts the augmented image $\mathbf{x}^+$ to the latent $\mathbf{z}_1$. We introduce noise interpolation $\mathbf{z}_\text{noise}$ into $\mathbf{z}_1$, following \citet{detok}, which is fed to the decoder head $\text{Dec}$ for reconstruction:
$$\mathcal{L}_{\text{rec}} = \mathcal{L}_{\text{L2}}(\text{Dec}(\mathbf{z}_\text{noise}), \mathbf{x}) + \mathcal{L}_{\text{LPIPS}}(\text{Dec}(\mathbf{z}_\text{noise}), \mathbf{x}) + \mathcal{L}_{\text{GAN}}.$$
Using $\mathbf{z}_\text{noise}$ as input provides latent space regularization (replacing the traditional KL-divergence term) and allows the $\mathcal{L}_{\text{rec}}$ and $\mathcal{L}_{\text{DSD}}$ objectives to share the same $\mathbf{z}_t$ input and a single forward pass through the ViT backbone.

\paragraph{Auxiliary Losses.}
We further enhance the learned features with alignment losses:
\begin{itemize}
    \item ViT Layer Alignment: Following REPA~\citep{repa}, we use cosine similarity to align the representations from the first few ViT layers (e.g., layers 0 and 1) to features from a strong teacher model like DINOv3~\citep{dinov3}.
    \item Representation-Level SD: The $\mathcal{L}_{\text{DSD}}$ primarily acts on the final latent space $\mathbf{z}$. We also introduce a self-distillation loss on the intermediate transformer representations to enforce broader feature-space alignment and stability.
    \item Classification Loss. ImageNet~\citep{imgnet} dataset have label for each image. We add a classification head with latent as input on and training classification with cross entropy loss. It helps model to build high quality latent.
\end{itemize}

\begin{figure*}[t]
    \begin{center}
    \includegraphics[width=1.0\linewidth]{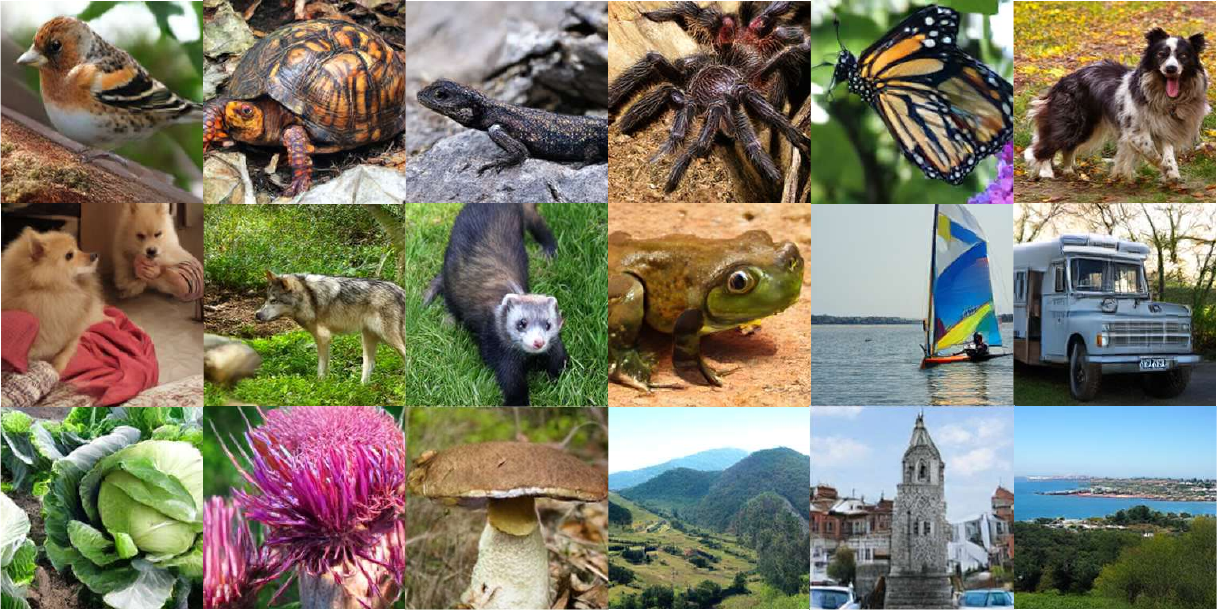}
    \vskip -0.05in
    \caption{Qualitative Results on Imagenet 256 $\times$ 256 using DSD-B. We use a classifier-free guidance scale $7.3$.}
    \label{fig:viz-results}
    \end{center}
    \vskip -0.2in
\end{figure*}

\begin{table*}[t]
    \centering
    \caption{System-Level Comparison between different generative models on ImageNet}\label{tab:sys}
    \begin{tabular}{lcccccccccc}
\toprule
Method & VAE & \multicolumn{2}{c}{Main Model}  &  \multicolumn{3}{c}{Generation w/o CFG} & \multicolumn{3}{c}{Generation w CFG}  \\ 
\midrule
~ & \#param & \#Epoch & \#param $\downarrow$ & gFID$\downarrow$ & sFID$\downarrow$ & IS$\uparrow$ & gFID$\downarrow$ & sFID$\downarrow$ & IS$\uparrow$ \\ 
\midrule
\rowcolor{blue!8}\multicolumn{10}{c}{AutoRegressive} \\
MaskGIT~\citep{maskgit} & - & 555 & 227M  & 6.18 & - & 182.1 & - & - & - \\ 
LlamaGen~\citep{llamagen} & 70M & 300 & 3.1B & 9.38 & 8.24 & 112.9 & 2.18 & 5.97 & 263.3 \\ 
VAR~\citep{var} & 84M & 350 & 2.0B &  - & - & - & 1.8 & - & 365.4 \\ 
MagViT-v2~\citep{magvitv2} & 230M & 1080 & 307M  & 3.65 & - & 200.5 & 1.78 & - & 319.4 \\ 
MAR~\citep{mar} & 66M & 800 & 945M  & 2.35 & - & 227.8 & 1.55 & - & 303.7 \\ 
MaskDiT~\citep{maskdit} & 84M & 1600 & 675M  & 5.69 & 10.34 & 177.9 & 2.28 & 5.67 & 276.6 \\ 
\rowcolor{blue!8}\multicolumn{10}{c}{Latent Diffusion} \\
DiT~\citep{dit} & 84M & 1400 & 675M  & 9.62 & 6.85 & 121.5 & 2.27 & 4.6 & 278.2 \\ 
SiT~\citep{sit} & 84M & 1400 & 675M  & 8.61 & 6.32 & 131.7 & 2.06 & 4.5 & 270.3 \\ 
FasterDiT~\citep{fasterdit} & 84M & 400 & 675M  & 7.91 & 5.45 & 131.3 & 2.03 & 4.63 & 264 \\ 
MDT~\citep{mdt} & 84M & 1300 & 675M  & 6.23 & 5.23 & 143 & 1.79 & 4.57 & 283 \\ 
MDTv2~\citep{mdtv2} & 84M & 1080 & 675M  & - & - & - & 1.58 & 4.52 & 314.7 \\ 
REPA~\citep{repa} & 84M & 80 & 675M  & 7.9 & 5.06 & 122.6 & - & - & - \\ 
~ & 84M & 800 & 675M  & 5.84 & 5.79 & 158.7 & 1.28 & 4.68 & 305.7 \\ 
REPA-E~\citep{REPA-E} & 70M & 40 & 675M  & 7.17 & 4.39 & 123.7 & - & - & - \\ 
& 70M & 80 & 675M  & 3.46 & 4.17 & 159.8 & 1.67 & 4.12 & 266.3 \\ 
~ & 70M & 800 & 675M  & 1.69 & 4.17 & 219.3 & 1.12 & 4.09 & 302.9 \\ 
LightningDiT~\citep{VAVAE} & 392M & 80 & 675M  & 4.29 & - & - & - & - & - \\ 
~ & 392M & 800 & 675M  & 2.05 & 4.37 & 207.7 & 1.25 & 4.15 & 295.3 \\ 
\midrule
DSD-S & - & 50 & 42M & 13.44 & 11.74 & 99.0 & 7.89 & 11.06 & 278.4 \\ 
DSD-M & - & 50 & 118M & 6.38 & 9.79 & 165.4 & 4.38 & 10.12 & 274.4 \\ 
DSD-B & - & 50 & 205M & 4.25 & 8.96 & 188.8 & 3.35 & 9.08 & 254.7 \\ 
\bottomrule
\end{tabular}%
\end{table*}

\section{Experiments}
\label{sec:exp}

Our foundational analysis of latent collapse and the proposed solution (DSD-D and DSD-T) has been empirically validated in the initial training steps, as demonstrated by the effective rank analysis in \cref{fig:erplot}.

We next validate the generative capability of our Diffusion Self-Distillation (DSD) framework. Specifically, we investigate two key research questions:
\begin{itemize}
\item How does DSD perform in image generation tasks compared with existing state-of-the-art generative models?
\item Does DSD's generative performance exhibit favorable scaling with model size, following established trends in large models?
\end{itemize}
Our code is available in the supplementary material.


\textbf{Implementation Details.}
We maintain the core setup of the LightningDiT framework~\citep{VAVAE} unless otherwise noted. All models are trained on the ImageNet~\cite{imgnet} training split. We adopt the standard data preprocessing protocol from ADM~\cite{adm}, where images are center-cropped and resized to $256 \times 256$ resolution. We use a standard ViT architecture~\citep{ViT} with a patch size of $16\times 16$ pixels, resulting in $256$ total spatial tokens per image. To enhance token representation, we utilize four additional register tokens, following the approach of \citet{ViTregister}.

For the compact latent space, we use $256$ tokens, each with a dimension of $16$ (i.e., a latent dimension of $d=16$). The specific heads for the unified ViT are:
\begin{itemize}
    \item Image Head (Encoder/Decoder): Composed of two convolution layers, similar to prior vision tokenizers~\citep{VAVAE, detok}.
    \item Diffusion Head: Composed of a single DiT layer~\citep{dit} and its final prediction head, used to predict both the clean latent ($\tilde{v}_\theta$) and the velocity ($v_\theta$).
\end{itemize}
For the auxiliary alignment Loss, we use a pretrained DINOv3-Base model~\citep{dinov3} as the external visual feature source. We apply the alignment objective (cosine similarity) specifically to the second layer of our DSD ViT backbone. 

In our experiments, DSD has three model sizes:
\begin{itemize}
    \item \textbf{DSD-S (Small):} 42 M parameters, 7 transformer layers, hidden dimension $512$.
    \item \textbf{DSD-M (Medium):} 118 M parameters, 11 transformer layers, hidden dimension $768$.
    \item \textbf{DSD-B (Big):} 205 M parameters, 15 transformer layers, hidden dimension $896$.
\end{itemize}

For optimization, we use the Muon optimizer~\citep{muon, KimiMuon}, with a weight decay of $1\text{e-}4$ and a base learning rate of $1\text{e-}4$. The batch size per GPU is set to $256, 224,$ and $192$ for DSD-S, DSD-M, and DSD-B, respectively. We apply gradient clipping with a global norm of $3.0$. The target encoder is updated using Exponential Moving Average (EMA) with a high rate of $0.99$. All experiments are conducted on 8 NVIDIA A800 GPUs using CUDA 12.8 and PyTorch 2.8~\citep{pytorch}. The wall-clock time per training epoch is approximately $14.0$, $19.9$, and $27.9$ minutes for DSD-S, DSD-M, and DSD-B, respectively.

\textbf{Evaluation.}
For image generation quality assessment, we strictly adhere to the established evaluation protocol used in ADM~\cite{adm}. We report the following standard metrics, measured on 50K generated images: Fréchet Inception Distance (FID)~\cite{fid}, Structural FID (sFID)~\cite{sfid}, Inception Score (IS)~\cite{is}. For sampling, we use the Euler sampler with 250 steps, following the LightningDiT approach~\citep{VAVAE}. 

\subsection{Image Generation Performance}

Visual examples of the generative output from our largest model, DSD-B, are provided in \cref{fig:viz-results}. The quantitative comparison against established models is presented in \cref{tab:sys}.

Our results demonstrate that DSD achieves outstanding generation quality across all metrics, especially in case without the use of classifier-free guidance ($\text{cfg}$). Notably, the DSD-B model, with only $205$ million parameters, significantly outperforms architectures with substantially larger parameter counts. For instance, DSD-B surpasses the generation FID achieved by the SiT model with REPA, despite the latter having a total parameter (VAE + Diffusion model) count exceeding $700$ million, and the tokenizer is each pretrain for a long time on larger OpenImage~\citep{openimg} dataset. This highlights the superior parameter efficiency of our unified, end-to-end trained architecture. Furthermore, the results clearly confirm the favorable scaling of the DSD framework. As the number of parameters increases from DSD-S to DSD-B, the generation performance improves significantly, confirming that the DSD design is a robust foundation for building large-scale generative models. 

\section{Conclusion}
In this work, we introduced Diffusion Self-Distillation (DSD), a novel framework that successfully unifies the VAE encoder, decoder, and diffusion backbone into a single, end-to-end trainable Vision Transformer (ViT), solving the prevalent efficiency and optimization issues of the decoupled LDM paradigm. The core technical hurdle, latent collapse, was resolved by a rigorous diagnosis revealing two failure modes in the standard diffusion loss: explicit suppression of latent variance and the violation of the rank differentiation mechanism due to the high-frequency nature of velocity prediction. Our solution stabilized training by (1) decoupling the target latent via a $\text{stop-gradient}$ operator, and (2) analytically proving the equivalence of velocity prediction to a clean latent expectation loss, which effectively forces the diffusion head to become a low-pass filter and successfully bridges diffusion modeling with stable self-distillation. Empirical results confirm DSD's superior efficiency and scalability, with the DSD-B model (only 205 M parameters) achieving a strong FID of 4.25 on ImageNet $256 \times 256$, demonstrating a powerful new direction for unified and parameter-efficient generative model design.

\section{Limitation}
Due to computation resource constraint, we do not scale DSD to larger model sizes aligned with our baselines. Moreover, we do not conduct experiments for verifying the effectiveness of our DSD as a unsupervised learning method. 

\newpage
{
    \small
    \bibliographystyle{ieeenat_fullname}
    \bibliography{main}
}

\end{document}